# Functional Co-Optimization Of Articulated Robots

Andrew Spielberg, Brandon Araki, Cynthia Sung, Russ Tedrake, and Daniela Rus

*Abstract*— We present parametric trajectory optimization, a method for simultaneously computing physical parameters, actuation requirements, and robot motions for more efficient robot designs. In this scheme, robot dimensions, masses, and other physical parameters are solved for concurrently with traditional motion planning variables, including dynamically consistent robot states, actuation inputs, and contact forces. Our method requires minimal user domain knowledge, requiring only a coarse guess of the target robot configuration sequence and a parameterized robot topology as input. We demonstrate our results on four simulated robots, one of which we physically fabricated in order to demonstrate physical consistency. We demonstrate that by optimizing robot body parameters alongside robot trajectories, motion planning problems which would otherwise be infeasible can be made feasible, and actuation requirements can be significantly reduced.

## I. INTRODUCTION

The design of robots and the design of robot motions are innately coupled problems. Changes in physical robot designs often require modifying or completely redesigning their motion primitives. Similarly, if trajectories that satisfy user constraints do not exist for a given robot in solving a required task, its mechanical design must be revisited.

Consider, for example, a Hexapod (Fig. 1) that must walk to and press a button $0.15$ m off the ground, while remaining planted on the ground itself. With leg lengths of $h = 0.10$ m, this is an impossible task as the Hexapod is not tall enough. However, lengthening the legs increases the torque effected by gravity on the leg joints, making it more difficult for the robot to walk. As the physical design changes, the motion and actuation for walking changes, *e.g.* legs with greater mass require more actuation; longer legs are able to traverse longer distances per gait cycle. Actuator selection needs to be considered as well. If smaller motors are used, link masses may need to be decreased in order to enable the desired robot motion; if larger motors are used, the robot chassis may need to grow to fit them.

Task-driven robot design typically involves several steps. Designers must begin by deciding, at a high level, how the robot will look and move, including what its method of movement and actuation will be. Following this, designers typically select actuators, settle on precise robot dimensions, materials, and other physical properties. When the robot has been fabricated, motions must be programmed and tested. This process can require many design iterations before the

A. Spielberg, B. Araki, C. Sung, R. Tedrake, and D. Rus are with MIT Computer Science And Artificial Intelligence Laboratory (CSAIL), Massachusetts Institute of Technology, 77 Massachusetts Avenue, Cambridge, Massachusetts, {aespielberg, araki, crsung, russt, rus}@csail.mit.edu.

This work was supported by The National Science Foundation Grant No. 1240383. We thank Michael Posa for discussions regarding the contact-implicit trajectory optimization method.

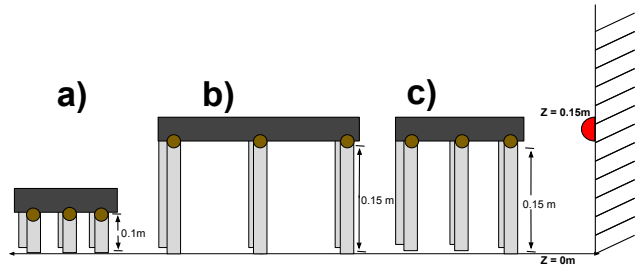

Fig. 1: A Hexapod robot must walk forward to hit a button on the wall. a) The Hexapod's legs are too short to reach the button. b) Lengthening the legs makes the Hexapod tall enough to reach the button c) Shortening the body allows for motions that are both feasible and require less energy.

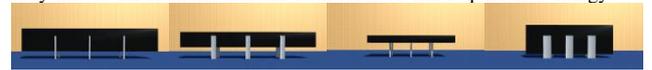

Fig. 2: Four parameterizations of the same Hexapod model. Though not visualized here, masses, moments of intertia, densities, and centers of mass of the links also are parameterized. More complex robots can have larger design spaces.

desired robot is built and able to complete the specified task.

We introduce a method for co-optimizing a robot's physical design and its motion. Our method operates on physically parameterized robot topologies (specifically, kinematic trees) with physical parameters such as link dimensions, masses, centers of mass, and moments of inertia. Such parameterized designs afford significant design space (see, *e.g.*, Fig. 2).

Our method is rooted in trajectory optimization [1], [2]. Trajectory optimization is a technique for finding kinodynamically feasible robot motions subject to constraints such as actuation limits, desired position and velocity states of the robot at certain times (*e.g.* starting pose), and environmental constraints, such as non-penetration with terrain or obstacles. Further, trajectory optimization methods optimize these feasible motions for specified objectives, such as minimizing energy expenditure, maximizing robot speed, and so on.

Traditional trajectory optimization methods typically operate on decision variables that include the robot state, actuation, and contact forces. We demonstrate that robot design parameters can be incorporated seamlessly into trajectory optimization problems, enabling the concurrent solution of robot trajectories, actuator selection, and physical designs.

We contribute the following:

- An algorithm for performing trajectory optimization with full dynamics while co-optimizing over physical design parameters and actuation requirements, with no prescribed gaits for walking robots.
- A system in which efficient evaluation of costs and constraints and their gradients and Jacobians respectively is achieved through symbolic representation of the robot's state, actuation, contacts, and parameters.
- Demonstrations of motions on a physical prototype

created using our approach.

*A. Prior Work*

Optimization techniques have shown great promise over recent years as a means of generating complex behaviors from higher level input. Work in [3] demonstrated how gaits could be designed interactively for walking robots. Work in [4] incorporated dynamics into the optimization and demonstrated that an online formulation of trajectory optimization could be used to create complex motions for humanoids by varying the constraints and dynamics. Similarly, [5] and [6] demonstrated that emergent behaviors can be synthesized for arbitrary morphologies from high level specification, and [7] demonstrated that trajectory optimization can be used to synthesize controllers for arbitrary morphologies and high-level gait specification. However, none of these works have incorporated physical parameter changes.

On the other hand, recent results related to design optimization include [8], where a kinematic approach to optimal linkage synthesis is proposed, [9], where robot function is specified in natural language to automate component selection, and [10], which demonstrates data-driven co-design for mechanical and electronic subsystems of functional robots. [11] demonstrated the ability to select discrete robot components, including batteries and actuators, based on constraints operating on mixed discrete and continuous variables, but does not consider geometry or motions.

The work most similar to ours is [12]. In this work, the authors demonstrate the potential for simultaneously optimizing robot gaits and design parameters. Their method does not factor in contact dynamics and allows only biologically-inspired subsets of potential quadrupedal foothold patterns. By contrast, our method for finding walking motions, which builds upon [1], does not rely on expert knowledge for gait patterns and computes contact forces implicitly as part of the optimization. Further, compared with their evolutionary algorithm-based approach, which provide few theoretical guarantees, we employ a gradient-based optimization approach that is locally optimal in costs and constraints.

Although the co-design of robot structure and motions has been gaining attention recently, it is worth noting that a related problem, the co-design of robot structure and robot *controllers* has been studied for far longer. For instance, a method for simultaneously designing a high-speed, dynamic arm with an optimal PD controller was explored in [13], and the simultaneous design of electric DC motors and their optimal PID controllers was explored in in [14].

## II. CO-DESIGN AS OPTIMIZATION

We frame co-design of robot structure, actuation, and motion as a parametric trajectory optimization problem, naturally folding in design parameters as additional variables atop the traditional trajectory optimization framework. By incorporating parameters directly in the optimization step, our method is able to alter the robot's physical structure to make otherwise infeasible tasks feasible, as well as optimize robots for tasks such as path following or forward walking. Our method exploits symbolic differentiation in order to compute gradients (Jacobians) of constraints with respect to physical parameters, enabling efficient optimization of these structural parameters. While we pay special interest to ground locomotion due to the complexity of the task, our method is general to a gamut of robot models and tasks.

Our algorithm solves for robot designs and fully dynamic motions. It ensures dynamical consistency by employing kinodynamic direct transcription trajectory optimization, solving for contact forces *via* implicit contact constraints as in [1]. Our algorithm operates in the continuous domain by solving for the robot state at discretized knot points and integrating the robot dynamics between them.

Together, these objectives and constraints combine to form a complex nonlinear program of the form:

$$\min_{\mathbf{h}, \mathbf{q}_1 \ldots \mathbf{q_k}, \dot{\mathbf{q}}_1 \ldots \dot{\mathbf{q}_k}, \lambda_1 \ldots \lambda_k, \mathbf{u}_1 \ldots \mathbf{u_k}, \rho} \mathcal{G} \quad (1)$$

subject to dynamics, contact constraints,

actuation constraints, parametric constraints,

design constraints, and task constraints where $\mathcal{G}$ is an objective function operating on the variables, and $\mathbf{q}$, $\lambda$ and $\mathbf{u}$ are robot state, contact forces, and actuation, respectively. We describe this model in more detail throughout the remainder of this section.

*A. Robot Specification and Dynamics*

Formally, let $R$ be a user-specified robot model comprised of links and joints with design parameters $\rho \in \mathbb{R}^p$. These design parameters may include, for each link, a parametric representation of its geometry and mass, as well as functions which define the link's inertial tensor, center of mass, and so on. Each joint may be actuated or non-actuated, and each actuator may possess torque limits.

Consider, for example, our Hexapod model. Here, $\rho$ includes as geometric variables the body length, width, and height and the length of each leg link, with the other dimensions fixed. $\rho$ also includes mass variables, namely the body mass and the leg mass (the same for each leg). The geometry of each link is a rectangular prism. We model each link as uniform density. We set the inertial tensor of each leg $L$ specifically to be:

$$\begin{pmatrix} m(w_L^2 + d_L^2)/12 & 0 & 0 \\ 0 & m(\mathcal{L}_L^2 + d_L^2)/12 & 0 \\ 0 & 0 & m(\mathcal{L}_L^2 + w_L^2)/12 \end{pmatrix}$$

where $\mathcal{L}_L$ and $m_L$ are parameters length and mass parameters of link $L$ respectively. The width $w_L$ and depth $d_L$ of link $L$ is constant in this case.

Key to our algorithm is paying special care to the effect of such parameters on robots' motion constraints. As dynamic constraints, we require:

$$\forall t, [\dot{\mathbf{q}}(t), \ddot{\mathbf{q}}(t)]^T = f(\mathbf{q}(t), \dot{\mathbf{q}}(t), \mathbf{u}(t), \lambda(t); \rho) \quad (2)$$

Here $f$ is a function governing the dynamics of a robot, and $\mathbf{q}(t) \in \mathbb{R}^n$, $\dot{\mathbf{q}}(t) \in \mathbb{R}^n$, $\mathbf{u}(t) \in \mathbb{R}^m$, $\lambda(t) \in \mathbb{R}^{6l}$, and $\rho \in \mathbb{R}^p$) are the robot state, time derivative of the robot state, actuation inputs (torques or forces), and contact forces at time $t$, and the design parameters, respectively. Parameters are a static attribute of the robot and are thus *not* a function of time.

By discretizing $t$ into a vector of $K$ (ordered) knot points $\mathbf{t} \in \mathbb{R}^K$, we can write, employing the backward Euler integration method:

$$\forall k \in 1 \ldots K, [\mathbf{q_{k+1}}, \dot{\mathbf{q}}_{\mathbf{k+1}}]^T = [\mathbf{q_k}, \dot{\mathbf{q}}_\mathbf{k}]^T + f(\mathbf{q_k}, \dot{\mathbf{q}}_\mathbf{k}; \mathbf{u_k}; \lambda_\mathbf{k}; \rho) dt_k$$
$$\text{where } dt_k = \mathbf{t_{k+1}} - \mathbf{t_k} \quad (3)$$

Again, note that $\rho$ is constant over *all* knot points. Specifically, for $f$, we compute the forward dynamics of the robot in order to compute the robot's manipulator equations and invert the equations in order to solve for $[\dot{\mathbf{q}}_\mathbf{k}, \ddot{\mathbf{q}}_\mathbf{k}]^T$. In other words, $f(\mathbf{q_k}, \mathbf{u_k}, \rho)$ is computed by solving the parameterized manipulator equations $\mathbf{H}(\mathbf{q_k}; \rho)\ddot{\mathbf{q}}_\mathbf{k} + \mathbf{C}(\mathbf{q_k}, \dot{\mathbf{q}}_\mathbf{k}; \rho)\dot{\mathbf{q}} + \mathbf{g}(\mathbf{q_k}; \rho) = \mathbf{B}(\mathbf{q_k})\mathbf{u} + \mathbf{J}\lambda_\mathbf{k}$ for $[\dot{\mathbf{q}}_\mathbf{k}, \ddot{\mathbf{q}}_\mathbf{k}]^T$. Here, $\mathbf{B}$ is the robot's control matrix, $\mathbf{J}$ is its contact point Jacobian matrix, $\mathbf{H}$ is its inertial matrix, $\mathbf{C}$ is its Coriolis matrix, and $\mathbf{g}$ are forces due to gravity. $\mathbf{H}$ is computed using the Composite-Rigid Body algorithm, and $\mathbf{C}$ is computed using the recursive Newton-Euler method.

For the contact forces, we add $\lambda = [\lambda_z, \lambda_x^-, \lambda_x^+, \lambda_y^+, \lambda_y^-, \gamma]$. Here, $\lambda_z$ is the normal for each of $l$ potential contact forces, $\lambda_x^+, \lambda_x^-, \lambda_y^+, \lambda_y^-$ are the bilateral sliding contact forces along the friction cone, and $\gamma$ is a slack variable in order to allow for sliding of the contacts. In practice, similar to [1], we add additional slack variables when solving the nonlinear complementarity constraints to make the problem better conditioned numerically.

For our Hexapod example, we model the center of each foot as being a potential contact point, giving $l = 6$. We then add constraints at each knot point $k$ as:

$$\phi(\mathbf{q_k}, \rho) \geq 0 \quad (4)$$
$$\lambda_{k,z}, \lambda_{k,x}^+, \lambda_{k,x}^-, \lambda_{k,y}^+, \lambda_{k,y}^-, \gamma_k \geq 0 \quad (5)$$
$$\mu\lambda_{k,z} - \lambda_{k,x}^+ - \lambda_{k,x}^- - \lambda_{x,y}^+ - \lambda_{k,x}^- \geq 0 \quad (6)$$
$$\phi(\mathbf{q_k}, \rho)^T \lambda_{k,z} = 0 \quad (7)$$
$$\gamma_k + \psi(\mathbf{q_k}, \dot{\mathbf{q}}_\mathbf{k}, \rho)^T \hat{\mathbf{e}}_x \geq 0 \quad (8)$$
$$\gamma_k - \psi(\mathbf{q_k}, \dot{\mathbf{q}}_\mathbf{k}, \rho)^T \hat{\mathbf{e}}_x \geq 0 \quad (9)$$
$$\gamma_k + \psi(\mathbf{q_k}, \dot{\mathbf{q}}_\mathbf{k}, \rho)^T \hat{\mathbf{e}}_y \geq 0 \quad (10)$$
$$\gamma_k - \psi(\mathbf{q_k}, \dot{\mathbf{q}}_\mathbf{k}, \rho)^T \hat{\mathbf{e}}_y \geq 0 \quad (11)$$
$$(\mu\lambda_{k,z} - \lambda_{k,x}^+ - \lambda_{k,x}^- - \lambda_{x,y}^+ - \lambda_{k,x}^-)\gamma_k \geq 0 \quad (12)$$
$$(\gamma_k + \psi(\mathbf{q_k}, \dot{\mathbf{q}}_\mathbf{k}, \rho)^T \hat{\mathbf{e}}_x)\lambda_{\mathbf{k},\mathbf{x}}^+ = 0 \quad (13)$$
$$(\gamma_k - \psi(\mathbf{q_k}, \dot{\mathbf{q}}_\mathbf{k}, \rho)^T \hat{\mathbf{e}}_x)\lambda_{\mathbf{k},\mathbf{x}}^- = 0 \quad (14)$$
$$(\gamma_k + \psi(\mathbf{q_k}, \dot{\mathbf{q}}_\mathbf{k}, \rho)^T \hat{\mathbf{e}}_y)\lambda_{\mathbf{k},\mathbf{y}}^+ = 0 \quad (15)$$
$$(\gamma_k - \psi(\mathbf{q_k}, \dot{\mathbf{q}}_\mathbf{k}, \rho)^T \hat{\mathbf{e}}_y)\lambda_{\mathbf{k},\mathbf{y}}^- = 0 \quad (16)$$

Here, $\phi_i(\cdot)$ is the signed distance between contact $i$ and the terrain, and $\psi_i(\cdot)$ provides the relative tangential velocity of each contact. (Note we have dropped the contact indexing $i$ in the constraints above for readability.) The first constraint encodes non-penetration; the second constraint is necessary for breaking the contact forces into slack and bilateral components. The third constraint enforces friction cone constraints, and the fourth constraint encodes the strict complementarity of contacts. The remaining constraints introduce the tangential velocity term to the previous constraints in order to allow for sliding in the contacts.

The decision to use an implicit contact constraint formulation is important for two reasons. First, it lessens the specification requirements for the user. As we are concerned with allowing users to design robots from high-level design goals, we consider such details to be a burden. Second, constraining foothold patterns can shrink the design space. This may make it impossible for the optimizer to find feasible designs and motions for a given task, even if feasible designs would otherwise exist; or, those solutions the optimizer finds may be highly suboptimal. By solving for contacts as we search for motions, we allow the optimizer to search over the space of all robot designs and motions at once.

*B. Constraints*

In addition to the robot, users may specify three types of problem constraints - parametric constraints, task constraints, and design constraints.

As a bare minimum for parametric constraints, we require an upper and lower bound on each design parameter in order to prevent degenerate solutions. If a user desires, auxiliary constraints may be added which establish a necessary relationship between parameters; these are useful for describing fabrication constraints. For example, the minimum possible mass of a robot may be a function of its size. For our Hexapod example, we require the (obvious) constraint that masses and dimensions must be positive, and specify an upper and a lower bound on each.

Task constraints, which define the robot's function, are kinematic in nature. These constraints allow users to specify the configuration (including world pose) of the robot, the position of any point *on* the robot in the world frame, and prescribe velocities of any of the robot's degrees of freedom. For walking robots, we also automatically add the constraint that links without contact points must not collide the ground (unless such a collision is explicitly requested by a user).

In our Hexapod task example, we add three task constraints: that our robot is upright on the ground with $\dot{\mathbf{q}}_\mathbf{1} = \mathbf{0}$ at the first knot point, that our robot is upright on the ground with $\dot{\mathbf{q}}_\mathbf{K} = \mathbf{0}$ at the final knot point, and that the front and center of the robot's body is at a height of $0.15$ m.

Design constraints define a mapping between actuation and design parameters. Higher power-output motors tend to be larger in both size and mass. We assume designers have access to a discrete collection of motors with which to build their robots. Each motor is rated with a maximum power output, and comes with its own design restrictions; specifically, they must fit on the robot in the specified locations, and the robot must be able to manage their load.

To ground this discussion, consider our Hexapod example. Our Hexapod topology calls for three actuators on each side. Any solution motion imposes constraints on the minimum motor size. The height of the Hexapod's body link must be at least the height of the motor; its width must be at least twice the width of the motor, and its length must be at least three times the length of the motor. Further, the mass of the body must be at least six times the mass of the chosen motor, plus the mass of the building materials.

Such a set of constraints is discontinuous and can cause numerical challenges for our optimizer. Therefore, we model the necessary lower bounds on mass and dimensions conservatively, linearly interpolating the motor masses and link dimensions between the jumps in power output. Although this constraint is nonsmooth at a few points, we have not found this to be an issue for our optimizer. Fig. 3 shows our model for conservatively estimating the added motor mass; similar models are used for the other variables as well.

As will be described in Section II-C, we seek to minimize the maximum power output throughout a motion, thus allowing our optimizer to choose motors with the smallest design restrictions possible. We currently require all motors in a robot to be the same for ease of programming the electronics.

### C. Objectives and Actuator Selection

Our algorithm can optimize over any smooth function of the robot state, actuation, contact, and design parameters. However, we mostly focus on the objective function

$$\mathcal{G} = \alpha \mathcal{G}_{act} + \beta \mathcal{G}_{reg} \qquad (17)$$

where $\mathcal{G}_{act}$ is an actuation minimizing term, $\mathcal{G}_{reg}$ is a parameter regularizer, and $\alpha$ and $\beta$ are user-specified weights.

By choosing $\mathcal{G}_{act}$ as part of our objective function, our method automatically chooses the actuators necessary for the designed robots by attempting to *minimize* the necessary power consumption. Since motors with higher power requirements are often larger, the decision to use higher power output motors constrains designs. We seek to minimize the motor size by minimizing the maximal power output.

We write this objective $\mathcal{G}_{act}$ as:

$$\mathcal{G}_{act} = \max_j \max_i \mathbf{u}_{ij} \equiv \|u\|_\infty \qquad (18)$$

where $\mathbf{u}_{ij}$ denotes the necessary actuation (torque) output of actuator $i$ and knot point $j$. The electrical power consumption of the motors we consider are linearly proportional to their torque applied, so we use torque as a proxy.

Through use of an auxiliary variable $\xi$, this objective can be written as:

$$\xi$$
$$\text{subject to } \forall i, j \quad \xi - \mathbf{u}_{ij} \geq 0$$
$$\xi + \mathbf{u}_{ij} \geq 0$$

The total actuation of each motor at each knot point must also be limited by the specifications of the largest motor available in a designer's library.

For aesthetics, functional purposes, and perhaps fabrication concerns (*e.g.* reducing material consumption), the designer may prefer that the output designs respect her vision as closely as possible. Therefore, we add the regularization term $\mathcal{G}_{reg}$ to the objective cost. We use $L_2$ regularization as its smoothness makes it easier to optimize over, though other regularizers may be employed as well. In particular,

$$\mathcal{G}_{reg} = \frac{1}{2} \|\rho - \rho_\mathbf{o}\|_2^2 \qquad (19)$$

The choice in $\alpha$ and $\beta$ represents a user-specified trade-off between the desire to minimize motor size and the desire to keep the design close a designer's original vision.

### D. Initialization

Finally, users must specify guesses for the robot's configuration $q$ at each time step and a maximum time for task completion. We stress here that the initial guess need not be physically feasible. Our algorithm generates initial poses for each knot point by linearly interpolating over a few sparse user-specified keyframes. An example five keyframe sequence for the Hexapod, inspired by the tripod gait, is shown in the video.

Velocities $\dot{\mathbf{q}}$ are automatically initialized to 0. Knot points are linearly spaced between $[0, T]$, where $T$ is a user-specified maximum duration of the motion. The contact force vector for contact point $i$ at knot point $k$ is set to be normal to the terrain with magnitude equal to the robot's weight if its keyframe interpolation is in contact with the terrain, and $\mathbf{0}$ otherwise. For actuation initialization, $\mathbf{u_{ij}} \sim \mathcal{U}(-u'_i, u'_i)$, where $u'_i$ is maximum torque-limit for actuator $i$.

### E. Optimization

A complete description of the algorithm can be found in Alg. 1. For the minimization step, we use SNOPT [15], a sparse sequential quadratic programming (SQP) solver which has been shown to be effective in solving direct transcription optimization problems due to their sparse nature. Our parameterized problem is in fact still sparse, since the number of design parameters in each constraint is constant and does not grow with the size of other problem parameters such as knot points or robot state dimensionality. SNOPT internally approximates the necessary Hessian information in the optimization step in a numerically robust manner. We break the optimization into two successive calls to SNOPT; once to find a feasible solution (with no objective function), and once using the feasible point as a seed to the objective optimization. Since SNOPT converts constraints to costs to find feasible points, minimizing over costs and constraints simultaneously can sometimes lead SNOPT to deem locally feasible problems as infeasible. By splitting the optimization into two steps, we avoid this scenario.

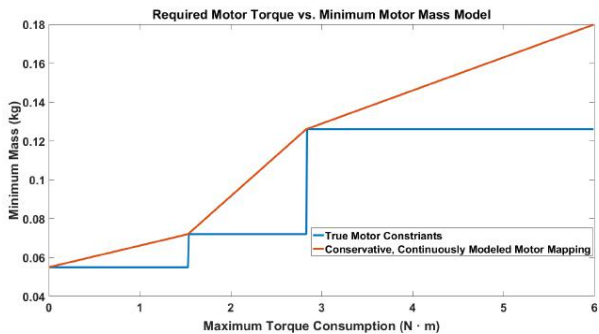

Fig. 3: A conservative mapping of the necessary output torque to individual minimum motor mass used to preserve continuity. We avoid flat regions in our model to ensure that our optimizer has meaningful gradients to work with.

## III. IMPLEMENTATION

Our system was implemented atop Drake [16], a MATLAB toolbox for robot simulation and trajectory optimization.

Trajectory optimization approaches often rely on gradient-based methods (such as SQP) to solve the formulated nonlinear programs. The form of these Jacobians is well understood for robot dynamics when taken with respect to actuation, state, and time. It is less obvious how to calculate these Jacobians with respect to arbitrary physical parameters. One approach is to calculate these Jacobians numerically. However, computing finite differences can be expensive and it can be numerically inaccurate in regions where function curvature is large.

Our solution is to represent costs and constraints symbolically in all decision variables. Since typical constraints vary smoothly with respect to physical parameters, we can use the symbolic representation to efficiently calculate these Jacobians. This has benefits for designer usability as well. A symbolic representation enables complex user costs and constraints with no user overhead in specifying Jacobians. This representation also makes it natural for users to define metaparameters of other physical parameters (*e.g.* moments of inertia that are functions of masses and link dimensions).

Generating symbolic objectives and constraints is a preprocessing procedure that only has to be performed once per robot or task. In order to speed up the evaluation of the constraints and their Jacobians, we compile them to C.

## IV. ANALYSIS

### A. Complexity

The algorithm has two phases of computation - preprocessing, in which dynamic, contact, and user-specified constraints and objectives are composed, and the optimization step itself. We examine each in turn.

*1) Constraints And Objectives:* The dynamics of the robot need only be computed once in order to formulate the dynamics constraints. This routine relies on two steps: Composite Rigid-Body and Recursive Newton-Euler algorithms, which run in $O(n)$ time, and solving the manipulator equations. Solving the manipulator equations is the most time-consuming operation, requiring $O(n^3)$ time. Computing the Jacobian of the dynamics requires differentiating each of the $O(n)$ dynamics expression coordinates with respect to $(n+m+p+l)$ state, actuation, contact force, and parameter variables, taking $O(n^3(n(n + m + p + l)))$ time.

Computation of the contact constraint relies on computing the distance from each contact point on the robot to each potential terrain contact point. Computation of each of the contact points in the world frame can be computed in $O(n)$ time using forward kinematics. Thus, the distance from each contact point in the world to each terrain point may be computed in $O(n)$ time. For piecewise defined terrain with $\tau$ grid cells, and $l$ contact points, $O(l\tau)$ constraints must be composed, this computation takes $O(l\tau n)$ time. For Jacobians, each of $O(l\tau)$ constraints must be differentiated with respect to $O(n+l+p)$ state, contact force, and parameter variables, taking $O(nl\tau(n + l + p))$ time.

Beyond the constraints that we add automatically, users are free to add their own constraints as described above.

*2) Optimization:* Each of our two SNOPT calls requires, at worst, $O(\|z_0-z^*\|_2)$ iterations to converge, where $z_0$ is the initial decision vector of each SNOPT call and $z^*$ is the local optimal feasible solution. For each iteration, every constraint and its Jacobian must be evaluated. The time needed to evaluate these constraints and Jacobians is equal to the time needed to formulate them, as described above. $O(K)$ dynamics, non-terrain penetration constraints, and contact constraint sets must be evaluated at each iteration.

### B. Convergence Guarantees

Our optimization problem is highly nonlinear over a very high dimensional space; state, design, and slack variables sum to $(2n+m+6l)K+p+1+(K-1)(5l+1)$ variables per problem. For nonlinear problems, SNOPT is only guaranteed to converge in the neighborhood of the solution. If the problem is initialized within a small enough neighborhood of a feasible solution, then SNOPT (and indeed our algorithm) is guaranteed to find a feasible design. Otherwise, no guarantee can be made. Theoretically, this neighborhood is very hard to define explicitly. In practice, we find that local feasibility of our problems is most sensitive to the initial guesses for the actuation. In the event of failures, random restart can eventually find a solution with most well-behaved problems.

## V. EXPERIMENTS

We present virtual experiments on four model robots to demonstrate the flexibility and power of our algorithm. Our model robots include a Hexapod, Biped, Quadruped, and Quadcopter, all with fully dynamic motion. Further results can be found in the video.

We model all of our virtual and physical experiments (with the exception of the Quadcopter) using the Dynamixel family of servo motors. A chart mapping the necessary power output to motors we have selected from the Dynamixel family, along with their masses and dimensions can be seen in Table I.

All walking robots were given 4 seconds to complete their tasks; the Quadcopter was given 6 seconds. $K = 16$ knot points were used for all experiments. For legged robots, we model the legs' inertial tensors as rectangular prisms as described in Section II-A; we choose constant values for the body links, assuming we can realize these moments by judiciously distributing the mass in our fabrication process. All joints on our robots are actuated, and there are no joint limits. Our robot models are shown in Figs. 4 and 5.

### A. Hexapod

Our Hexapod demonstrates our running example being optimized using our algorithm. We require the Hexapod to walk forward $0.45$ m and touch a height of $0.15$ m, while keeping all legs planted on the ground. This task is infeasible without adjusting its parameters, since the center of its body initially is only $0.1$ m tall. The Hexapod has 6 parameters in total: the $X$ (length), $Y$ (width), and $Z$ (height) extents of the base, the (equal) height of the legs, the mass of the body,

## Algorithm 1 Co-optimization Of Trajectories and Parameters

1: **procedure** CO-OPTIMIZE($R, U, \mathcal{G}$)  ▷ $R$ is a parameterized robot design, $\mathcal{G}$ is a user-specified objective function, $U$ is the set of user constraints, $\mathcal{G}$ is a user-specified objective function.
2:     Constraints ← $\{U\}$
3:     Costs ← $\{\}$.
4:     **for** $k = 0 \ldots K - 1$ **do**
5:         Constraints ← Constraints $\cup \{\Lambda_k, M_k, \mathbf{q}_{k+1} = \mathbf{q}_k + f(\mathbf{q}_k, \mathbf{u}_k, \rho)dt_k, \rho_{lower} \leq \rho \leq \rho_{upper}\}$
6:                                                                       ▷ $\Lambda_k$ are the contact constraints, $M_k$ are the motor constraints.
7:     $(q, , \dot{q}, u, dt, \lambda, \rho) \leftarrow$ SNOPT$(q_{init}, \dot{q}_{init}, u_{init}, dt_{init}, \lambda_{init}, \rho_{init},$ Constraints, Costs$)$  ▷ Solve for feasibility.
8:     Costs ← $\mathcal{G}$
9:     **return** (Success, $q, \dot{q}, u, dt, \lambda, \rho$) ← SNOPT$(q, \dot{q}, u, dt, \lambda, \rho,$ Constraints, Costs$)$  ▷ Solve with costs.

| Motor Name | X Extents (m) | Y Extents (m) | Z Extents (m) | Mass (kg) | Torque (N · m) |
|---|---|---|---|---|---|
| AX-12a | 0.032 | 0.04 | 0.05 | 0.055 | 1.53 |
| RX-28 | 0.0355 | 0.0356* | 0.41 | 0.072 | 2.83 |
| MX-64T8 | 0.0402 | 0.041 | 0.0611 | 0.126 | 6.0 |

TABLE I: The motors from our selected Dynamixel motor library, ordered by increasing torque and size. All motors consume 12 V and have a top speed of approximately $2\pi$ rad / s.
*Though this motor's $Y$ extents are 0.0356, we model it as 0.0405 in order to keep the Y extents monotonically increasing.

and the (equal) mass of the legs. The optimization problem has 1528 decision variables.

We use the Hexapod as an opportunity to demonstrate the results of our regularization objective. In particular, we run our Hexapod under two conditions: $\beta = 0$, and $\beta = 1.0$. We set $\alpha = 1.0$ for all of our experiments. For each condition, we ran 10 optimizations. Initial parameters and the results under each condition can be seen in Table II.

As expected, the legs of the Hexapod grow to satisfy the height constraints. Further, the leg masses decrease, reducing the power output needed to propel the robot forward.

As $\beta$ increases, the Hexapod approaches its initial parameter values where possible. Larger $\beta$ results in larger link masses; larger motors may be necessary depending on the trajectories selected. For the $\beta = 0$ case, the motors had to be increased from the AX-12a model to the RX-28 in one trial, and in the $\beta = 1.0$ case, the motors had to be increased in four of the trials. As expected, larger relative $\beta$ values decrease the emphasis on reducing motor size.

We found in our $\beta = 0$ experiments that some of the optimized parameter values were spread fairly uniformly throughout their range; for larger $\beta$ values these parameters snap to their initial parameter values (*e.g.* the body $Z$ extents). The regularizer can automatically detect parameters which have little effect on the actuation objective and keep them in their preferred configuration with little to no cost.

We fabricated a real Hexapod in order to demonstrate that our simulations are physically realizable. We selected one optimized Hexapod design from our $\beta = 0$ experiments and fabricated it with a light laser cut acrylic frame, 3-D printed legs, and 6 Dynamixel AX-12a servos with PID position and velocity control. An Arduino MEGA 2560 was used to control the electronics and record data. Our fabricated Hexapod body is 0.9 m long, 0.1 m wide, and 0.15 m tall. Its legs are 0.15 m tall legs.

A montage of one fabricated Hexapod alongside its simulated counterpart can be found in Fig. 4. Compared with its planned motion, the center of mass of the fabricated Hexapod had an average integrated squared error of 1.93 cm$^2$ in $x$ (the direction of motion), 0.078 cm$^2$ in the $y - z$ plane, and $1.97^{\circ^2}$ in its angular orientation. Although geometrically the motions look quite similar, there is a large discrepency in tracking in the $x$ direction since the actual friction coefficients were higher than those used for planning the virtual model. Throughout its trajectory, the center of mass of the physical model gradually fell behind that of the virtual model due to slip. Improving the accuracy of the frictional coefficients or tightly constraining the allowed slip in motion planning can reduce these types of errors.

*B. Biped*

The Biped example demonstrates that our walking robots can locomote on more than just flat terrain. In this example, we present a Biped, consisting of a base, hip joints, and knee joints. It consists of the following eight parameters: the length of the lower leg links (equal), the length of the upper leg links (equal), the $x$, $y$, and $z$ dimensions of the base, the mass of the lower leg links (equal), the mass of the upper leg links (equal), and the mass of the body.

We present two tasks for the Biped. In the first task, we require the Biped to walk forward 0.3 m. In the second task, we require the Biped to walk forward 0.3 m and up a 0.1 m step. We set $\alpha = 1.0$ and $\beta = 0$ for both tasks. Further, we restrict the final height of the Biped to lie between 0.25 m and 0.35 m tall. This problem has 750 decision variables.

We ran 30 optimization runs for each task. Table III presents statistics about the optimized design and motion, along with valid parameter ranges. Two items are of particular note. First, resulting masses are near their lower limits. Lowering the masses of the links decreases the power needed to actuate them, and allows the optimizer to select the smallest motors available (which it does for all tasks). Notably, the masses are rarely all at their absolute minimum - some excess mass may be added by the optimizer to increase the controllability of the design. Second, leg links tend to get longer, meaning they have to actuate over a smaller angular range. For the step task in particular, the algorithm discovers that the upper leg link should be shorter than in the flat

| $\beta$ | Actuation Objective (N · m) | Regularization Objective | Body X (m) 0.9 / 1.2 / 1.05 | Body Y (m) 0.05 / 0.10 / 0.10 | Body Z (m) 0.1 / 0.15 / 0.15 | Leg Mass (kg) 0.08 / 0.12 / 0.10 | Body Mass (kg) 0.3 / 0.5 / 0.4 | Opt. Time (s) |
|---|---|---|---|---|---|---|---|---|
| 0 | 1.49 (0.572) | 0.394 (0.019) | 1.01 (0.154) | 0.092 (0.008) | 0.138 (0.023) | 0.08 (0.0) | 0.027 (0.46) | 2,246 (2,230) |
| 1.0 | 1.28 (0.189) | 0.447 (0.026) | 1.50 (0.014) | 0.1 (0.0) | 0.15 (0.0) | 0.08 (0.0) | 0.433 (0.019) | 2,049 (2,007) |

TABLE II: Results of the simulated Hexapod optimizations for two $\beta$ values. Parentheses denote standard deviations. Headers of parameter columns denote lower bound / upper bound / initial value. The task was feasible for 7 of the 10 initializations.

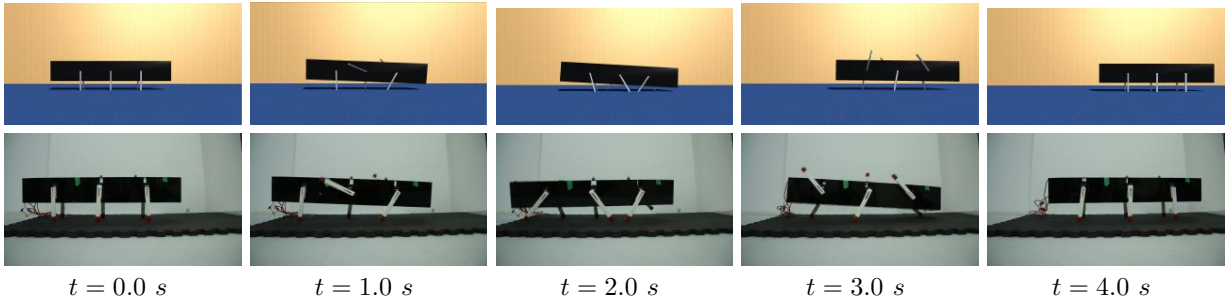

$t = 0.0\ s$     $t = 1.0\ s$     $t = 2.0\ s$     $t = 3.0\ s$     $t = 4.0\ s$

Fig. 4: A montage of our simulated fabricated Hexapod in motion.

| Task | Actuation Objective (N · m) | Lower Leg Length (m) 0.5 / 4.0 / 1.0 | Upper Leg Length (m) 0.5 / 4.0 / 2.0 |
|---|---|---|---|
| Flat | 0.506 (0.117) | 0.216 (0.044) | 0.100 (0.032) |
| Step | 0.651 (0.267) | 0.2328 (0.509) | 0.085 (0.044) |

| Task | Body X (m) 0.3 / 2.0 / 1.0 | Body Y (m) 0.3 / 2.0 / 1.0 | Body Z (m) 0.3 / 2.0 / 1.0 | Body Mass (kg) 0.5 / 10.0 / 4.0 |
|---|---|---|---|---|
| Flat | 0.080 (0.00) | 0.085 (0.079) | 0.133 (0.044) | 0.14 (0.058) |
| Step | 0.088 (0.024) | 0.097 (0.076) | 0.130 (0.043) | 0.126 (0.026) |

| Task | Upper Leg Mass (kg) 0.5 / 10.0 / 4.0 | Lower Leg Mass (kg) 0.5 / 10.0 / 4.0 | Opt. Time (s) |
|---|---|---|---|
| Flat | 0.064 (0.037) | 0.068 (0.030) | 685 (1,830) |
| Step | 0.066 (0.034) | 0.059 (0.019) | 460 (500.2) |

TABLE III: Results of the simulated Biped optimizations for the flat and step terrain. Parentheticals denote standard deviations. Headers of parameter columns denote lower bound / upper bound / initial value.

walking task in order to clear the step *via* a forward motion.

### C. Quadruped

The Quadruped presents the most articulated robot of our examples. Our Quadruped model consists of four legs attached to a base. Similar to our Biped, each leg has an upper and lower limb; the parameterization used for the Quadruped, in fact, is the same as that of the Biped but with two extra legs (and some different parameter ranges). We require the robot to walk forward $0.25$ m, with a final height between $0.35$ m and $0.4$ m. Note the Quadruped only starts at $3$ m tall; the algorithm must discover that it must grow. The Quadruped problem has $1284$ decision variables.

Although we have put a focus on actuation minimization and regularization objectives, our method can extend to other objectives which may be of interest to other roboticists. To demonstrate this, we optimize our Quadruped over two sets of experiments. First, we optimize our Quadruped over our standard objectives with $\alpha = 1.0$ and $\beta = 0.0$. Second, we optimize our Quadruped in a different set of experiments where we minimize time to task completion.

Each experiment was run $10$ times. This is a difficult and time-consuming optimization problem, sensitive to initial actuation guesses, and so for each experiment we only received a handful of successes (three each). Thus, we note mostly qualitative results here.

In the actuation minimization experiments, the heights of the legs increased to their maximum combined lengths of $0.4$ m. This allows the robot to locomote while expending as little torque possible, since its joints have to cover less of an angular range to travel the required distance. Meanwhile, in the velocity tasks, the legs decreased to their minimum combined lengths of $0.35$ m. Shorter legs give the robot greater acceleration and deceleration, allowing the robot to achieve the zero velocity start and end constraints.

In both sets of experiments, the base and upper leg link masses decreased to nearly their minimum bounds; the total robot masses decreased overall as well. However, the bottom link mass decreased for the actuation minimization task, while increasing slightly for the time minimization. Increasing mass in the lower leg links increases their inertial moments, making overall velocity of the robot easier to control. This is again important for allowing time-objective designs to easily decelerate. Controllability is not such a concern for the lower-actuated design, and so the optimizer is free to decrease masses to decrease power output.

Finally, we notice a large contrast in the objective values between the two tasks. For the actuation minimization task, the full $4$ seconds was required, while the maximum torque required ranged between $0.294$ and $0.494$ N · m, resulting in the smallest motors. Meanwhile, for all time minimization trials, the largest motors were selected and the maximum torque was used ($6$ N · m), with most of these high torques occurring during the push-off and landing of contacts. Times required for task completion, however, ranged from $1.252$ s to $1.268$ s, marking a significant improvement.

### D. Quadcopter

As a final example, we present a Quadcopter, which we use to validate our assertion that our algorithm is able to find known optimal parameterizations, with better objectives than any other (valid) fixed parameterization. The Quadcopter also presents an opportunity to demonstrate our algorithm on a non-walking example, demonstrating its generalizability.

Our Quadcopter consists of a circular base with motors spaced evenly around the circle half the overall size of the Quadcopter. We model the base as being uniform density with fixed inertia. Each motor acts as a force element which can provide an impulsive linear force, which provides linear

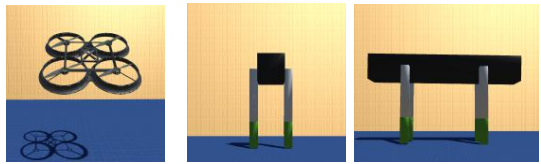

Quadcopter model    Biped model    Quadruped model

Fig. 5: The non-Hexapod models.

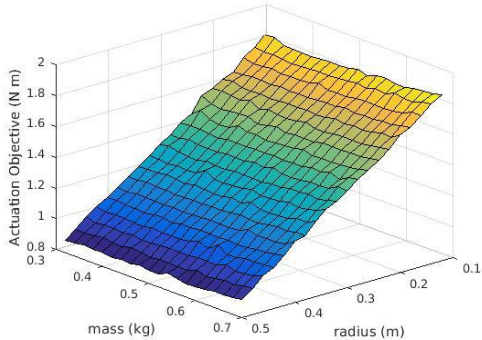

Fig. 6: A surface plot of the actuation costs versus various parameterizations for the Quadcopter. Our parameterized trajectory optimization is successfully able to find the optimal parameterization at the bottom-left corner.

thrusts and torques to the Quadcopter's base. Our Quadcopter has two parameters - the radius of the base (which in turn positions the motors), and an adjustable mass which can be increased at each motor's base. We consider the motors chosen *a priori* and massless for this single example.

For our task, we require the Quadcopter to fly in a circle of radius 1 m by hitting 16 waypoints. This simple model and task (274 decision variables) admits an obvious optimal solution: in order to maneuver with as little power output as possible, torquing the robot and maintaining altitude must be as cheap as possible. This is achieved when the motors are extended as far out radially as possible and when the mass is reduced to its minimum.

In our experiments, we allow the Quadcopter radius to vary between 0.1 m and 0.5 m (initialized at 0.3 m,) and allow the mass to vary between 0.3 kg and 0.7 kg (initialized at 0.5 kg). Over 20 trials, our optimizer successfully found the optimal configuration every time in less than 5 seconds. Beyond this, we ran 20 trials fixed at 400 different parameter configurations. A plot showing the resulting average costs of these trials can be seen in Fig. 6. The minimum can be found at 0.5 m and 0.3 kg as claimed; our algorithm recovered the optimal parameterization and corresponding motion.

## VI. CONCLUSION

We have introduced an algorithm for the co-optimization of robot design and function, and we demonstrate that a natural parametric representation and complete symbolic formulation of constraints, costs, and robot dynamics allows for direct and efficient of physically realizable designs. Our method is able to simultaneously design the robot motion, robot body parameters, and select from available actuators with very little user overhead. Our work suggests two natural follow-up problems whose solutions would aid the effectiveness and usefulness of our method.

First, our method would benefit from a better initialization procedure. Our method assumes that the robot designer can identify initial guesses for the robot motions and can specify them as waypoints. In our examples these initial guesses were fairly obvious, typically amounting to linear interpolation between starting and ending poses and considering natural biological motion of legs. Further, our method chooses the initial actuation vector at random, which can sometimes lead to local infeasibilities, requiring a random restart. In order to automate the design process further, a similar coarse planning method such as a parametric variant of RRT [17] is needed.

Second, topology is fixed in our method. A method for searching over morphology while searching over design parameters is necessary to explore larger design spaces. Such an optimization would likely be a mixture of discrete and continuous variables, and thus much harder to solve efficiently, though we have already started in this direction using conservative continuous models for motor selection.